\documentclass[11pt]{article}

\usepackage{longtable,booktabs,array,tabularx,float}
\usepackage{placeins}

\usepackage{acl}
\usepackage{microtype}
\usepackage{stfloats}

\makeatletter
\renewcommand{\cite}[1]{\hyperlink{citeproc-#1}{\citeproctext}}
\renewcommand{\bibitem}[2][]{\hypertarget{citeproc-#2}{}\item\ignorespaces}
\def\@biblabel#1{}
\makeatother

\usepackage{fontspec}
\newfontfamily\cjktext{bkai00mp.ttf}

\usepackage{graphicx}
\usepackage{amsmath,amssymb}

\providecommand{\tightlist}{\setlength{\itemsep}{0pt}\setlength{\parskip}{0pt}}
\usepackage{xcolor}
\newcommand*{\citeproctext}{}
\newcommand*{\citeproc}[2]{\begingroup\def\citeproctext{#2}\cite{#1}\endgroup}
\makeatletter
\let\@cite@ofmt\@firstofone
\def\@biblabel#1{}
\def\@cite#1#2{{#1\if@tempswa , #2\fi}}
\makeatother
\newlength{\cslhangindent}
\newlength{\csllabelwidth}
\newenvironment{CSLReferences}[2]
 {\begin{list}{}{%
  \setlength{\itemindent}{0pt}
  \setlength{\leftmargin}{0pt}
  \setlength{\parsep}{0pt}
  \ifodd #1
   \setlength{\leftmargin}{\cslhangindent}
   \setlength{\itemindent}{-1\cslhangindent}
  \fi
  \setlength{\itemsep}{#2\baselineskip}}}
 {\end{list}}
\usepackage{calc}

\title{Inside the Scorer: Mechanistic Evidence on Rubric Control in
LLM-Based Automated Essay Scoring from 25 Base Models}

\author{Chungche Huang {\cjktext 黃崇哲} \\ Graduate Institute of Linguistics \\ National Taiwan University \And Po-Ya Angela Wang {\cjktext 王伯雅} \\ Co-intelligence Humanities AI \\ (CHAI) Future Lab \\ National Taiwan University \AND Shu-Kai Hsieh {\cjktext 謝舒凱} \\ Graduate Institute of Linguistics \\ National Taiwan University \AND \texttt{\{chungchehuang, poyaw, shukaihsieh\}@ntu.edu.tw} \\ {\small\itshape Accepted at PACLIC 40, Taipei, Taiwan, December 10--12, 2026}}

\begin{document}
\maketitle

\begin{abstract}
Automated essay scoring (AES) using large language models (LLMs) offers
promising use cases for language assessment, yet high-stakes testing
requires scores that are both controllable and verifiable, which current
LLM-based scorers lack. Given that, we investigate whether LLM-based
scorers form distinct internal representations for individual scoring
rubrics and whether these representations can be independently
controlled. Probing and steering 25 open-weight base LLMs (0.13B--10.31B
parameters; 14 families) on human-scored BESTEP English writing tasks,
we find that holistic scores are decodable from model activations for
the longer tasks (Parts 2 and 3) and only weakly for the shortest (Part
1). Furthermore, scoring rubrics have partly separable directions in
representation space for the longer tasks, though not for short
responses (Part 1). However, causal steering reveals a critical
limitation: applying a matching intervention (e.g., steering ``grammar''
on a grammar-degraded response) yields only a negligible advantage
(about 2\% of the effect) over a mismatched intervention (e.g., steering
``vocabulary''), and none that is detectable in any of the 10 testable
comparisons across reliably measured rubrics. Rather than controlling
distinct evaluative dimensions, the rubric directions share a dominant
lever whose rubric-specific parts are not selective, which may mirror
the halo structure of human ratings. Our contributions are two-fold.
First, we show that analytic scoring rubrics designed for human raters
do not behave as independent controls in base LLMs, even among the seven
models whose scores agree best with human raters (\emph{r} ≥ 0.6).
Second, we establish a methodology that shows that score agreement and
decodability do not imply rubric-level control in LLM-based assessment.
\end{abstract}

\emph{Keywords:} Automated Essay Scoring, Large Language Models,
Rubric-Based Assessment, Interpretability, Rubric Representations

\section{Introduction}\label{introduction}

Automated essay scoring (AES) has recently started implementing neural
network models, e.g., large language models (LLMs), in the grading
process
(\citeproc{ref-Johnsi2026-ComprehensiveSurveyMethodologies}{Johnsi and
Kumar, 2026}). Agreement with human raters rose along the way, but the
models remain opaque to inspection. Transparency and interpretability
remain open priorities in AES investigations
(\citeproc{ref-Johnsi2026-ComprehensiveSurveyMethodologies}{Johnsi and
Kumar, 2026}). Control and verifiability are hard to explain in
LLM-based AES systems, which is one of the biggest challenges when
implementing LLM technologies in high-stakes language assessments. Test
providers respond in different ways. Some rely on mixed models where
human raters have the final score. Some develop feature-based machine
learning models in which humans define what is measured. Others, such as
the Duolingo English Test (DET), decompose the writing construct into
subconstructs and constrain any LLM-derived feature score to a single
subconstruct, so that the model cannot influence the holistic score
except through that one slot. As stated in the report, scoring directly
with deep learning ``cedes both control and verifiability''
(\citeproc{ref-Madnani2026-ScoringWritingSpeaking}{Madnani et al.,
2026}).

Although AES designs such as the DET minimize the black box impact on
the score, the core question remains unsolved: what exactly an LLM
scorer does on the inside. As LLMs become more and more ubiquitous in
language assessment, we need to know whether a model's internal
representations contain the scoring rubrics at all, and whether each
scoring rubric is controllable and verifiable separately. Current score
agreement benchmark tests cannot answer this. Prior interpretability
work on AES examines what a scorer has learned or which features it
weights
(\citeproc{ref-Fiacco2023-ExtractingUnderstandingImplicit}{Fiacco et
al., 2023}; \citeproc{ref-Wang2026-OpeningBlackboxLLMbased}{Wang et al.,
2026}). We approach the topic with two AES principles in mind, control
and verifiability, by adding causal intervention on the representations
themselves, mirroring how these systems operate in real-world
assessments.

In this study, we mechanistically audit LLM-based AES by probing and
steering 25 open-weight base LLMs (0.13B to 10.31B parameters, 14
families) as they score BESTEP writing responses across three task types
and six analytic scoring rubrics (see section Setup). We carry out the
investigation guided by three questions. RQ1: Is the human holistic
score decodable from a model's internal activations, and where does that
signal live? Which layers have the clearest scoring signals? RQ2: Do the
six scoring rubrics occupy distinguishable directions in representation
space? RQ3: Does steering a rubric's direction move the model's score
specifically for that scoring rubric, or does any rubric-adjacent push
move it about equally? In this paper, we choose base models as a
baseline, so that fine-tuned variants (instruction-tuned, chat) can be
compared against them in future studies. Our contributions are two-fold.
First, we establish a white-box audit protocol for LLM-based automated
essay scoring (AES) that disentangles a rubric's decodability,
separability, and causal control. Moreover, we show that analytic
scoring rubrics designed for human raters do not behave as independent
controls in base LLMs: for the rubrics that steer the score and are
reliably measured, steering one rubric's direction moves the score about
as much as steering a different one, even among the models whose scores
agree best with human raters. The code and the database behind every
number are public for verification at
\url{https://github.com/lopentu/inside-the-scorer-paclic2026}.

\section{Related Work}\label{related-work}

\textbf{AES and LLM-based scoring.} As surveyed in Johnsi and Kumar
(\citeproc{ref-Johnsi2026-ComprehensiveSurveyMethodologies}{2026}), AES
first moved from feature-engineered machine learning methods to
BERT-based deep learning, e.g., through domain-adaptive training,
multi-scale essay representations, and hierarchical transfer learning
(\citeproc{ref-Ramesh2022-AnAutomatedEssayScoringSystem}{Ramesh and
Sanampudi, 2022}; \citeproc{ref-Wang2022-UseBertAutomated}{Wang et al.,
2022}; \citeproc{ref-Cao2020-DomainAdaptiveNeuralAutomated}{Cao et al.,
2020}; \citeproc{ref-Xue2021-HierarchicalBERTBasedTransfer}{Xue et al.,
2021}; \citeproc{ref-Settles2020-MachineLearningDriven}{Settles et al.,
2020}). Additionally, a second wave of AES investigations focuses on
LLMs as scorers, resulting in mixed findings. Some studies report LLM
ratings holding up under real psychometric scrutiny
(generalizability-theory and many-facet Rasch analyses) across thousands
of essays and multiple genres
(\citeproc{ref-Song2025-ExploringLLMAutoscoringb}{Song et al., 2025};
\citeproc{ref-Wang2025-EvaluatingLargeLanguage}{Wang et al., 2025a}),
and case studies where an LLM produces both a score and trait-level
comments across several analytic scoring rubrics at once
(\citeproc{ref-Wang2025-LLMsCanPerform}{Wang et al., 2025b};
\citeproc{ref-Cho2024-DualscaleBERTUsing}{Cho et al., 2024}). Others
find the opposite: robustness failures under adversarial stress-testing
and demographic bias in ratings
(\citeproc{ref-Kabra2022-EvaluationToolkitRobustness}{Kabra et al.,
2022}; \citeproc{ref-Yamashita2025-ExploringPotentialBiases}{Yamashita,
2025}), systematic disagreement between LLM and human grading behavior
(\citeproc{ref-Mathew2026-LLMsNotGrade}{Mathew et al., 2026}), and
validity that shifts by language, genre, and proficiency level rather
than holding uniformly
(\citeproc{ref-Mizumoto2023-ExploringPotentialUsing}{Mizumoto and
Eguchi, 2023}; \citeproc{ref-Pack2024-LargeLanguageModels}{Pack et al.,
2024}; \citeproc{ref-Mansour2024-CanLargeLanguage}{Mansour et al.,
2024}; \citeproc{ref-Gaggioli2025-AssessingReliabilityValidity}{Gaggioli
et al., 2025}; \citeproc{ref-Barbosa2025-EmpiricalAnalysisLarge}{Barbosa
et al., 2025}; \citeproc{ref-Yavuz2025-UtilizingLargeLanguage}{Yavuz et
al., 2025}; \citeproc{ref-Yancey2023-RatingShortL2}{Yancey et al.,
2023}). A related strand looks past scoring accuracy to classroom use,
from ChatGPT's promise and pitfalls as a writing tutor to its measurable
effect on students' own writing when used for feedback
(\citeproc{ref-Barrot2023-UsingChatGPTSecond}{Barrot, 2023};
\citeproc{ref-Ding2025-ChatGPTAutomatedWriting}{Ding et al., 2025}).

\smallskip

\textbf{Control and verifiability in operational scoring.} High-stakes
providers mostly manage the opacity of LLM-based AES systems through
design. The Duolingo English Test decomposes a subconstruct into
features and uses various LLMs and machine learning models across
subconstructs and grader versions. Thus, an LLM-derived feature rating
contributes to the score only through its own subconstruct, and the
contribution of each subconstruct can be measured and explained
(\citeproc{ref-Madnani2026-ScoringWritingSpeaking}{Madnani et al.,
2026}). The same report notes that it remains substantially harder to
determine why an end-to-end system assigned a particular score than it
is for feature-based systems.

Prior analyses of neural network AES scorers are observational. Wang et
al. (\citeproc{ref-Wang2026-OpeningBlackboxLLMbased}{2026}) analyze how
LLM scorers weight textual features, and Fiacco et al.
(\citeproc{ref-Fiacco2023-ExtractingUnderstandingImplicit}{2023}) align
components of a transformer scorer's activations with
human-understandable feature groups. In our study, we approach this
topic with high-stakes language assessment's grading principles, i.e.,
control and verifiability, and intervene causally on the rubric
representations.

A separate line augments scores with rubric-level explanations or
rationales (\citeproc{ref-Bueno2026-ScoringExplanationsEvaluating}{Bueno
et al., 2026}; \citeproc{ref-Chu2025-RationaleEssayScores}{Chu et al.,
2025}; \citeproc{ref-Jiao2025-ExploringUtilitiesRationales}{Jiao et al.,
2025}; \citeproc{ref-Kucia2026-LLMEssayScoring}{Kucia et al., 2026}).
These works evaluate rubric outputs, whereas we evaluate internal rubric
representations in LLM-based AES.

\smallskip

\textbf{Mechanistic interpretability and faithfulness.} Our probing and
steering methodology draws on activation patching
(\citeproc{ref-Zhang2023-BestPracticesActivation}{Zhang and Nanda,
2023}), causal concept-based explanations
(\citeproc{ref-Wu2023-CausalProxyModels}{Wu et al., 2023}), and prior
work examining whether model-generated explanations faithfully reflect
internal computations
(\citeproc{ref-Atanasova2023-FaithfulnessTestsNatural}{Atanasova et al.,
2023}; \citeproc{ref-Jacovi2021-AligningFaithfulInterpretations}{Jacovi
and Goldberg, 2021};
\citeproc{ref-Chuang2026-FaithLMFaithfulExplanations}{Chuang et al.,
2026}; \citeproc{ref-Ferrando2023-ExplainingHowTransformers}{Ferrando et
al., 2023};
\citeproc{ref-Demircan2024-SparseAutoencodersReveal}{Demircan et al.,
2024}). This concern applies to AES, because a plausible-sounding
rationale does not guarantee that the model actually exercises
rubric-specific control.

Our two evaluation checks are grounded in probing research. First,
Elazar et al. (\citeproc{ref-Elazar2021-AmnesicProbingBehavioral}{2021})
demonstrate that a property's being decodable from a representation does
not mean the model actually uses it. Second, Hewitt and Liang
(\citeproc{ref-Hewitt2019-DesigningInterpretingProbes}{2019}) show that
probe accuracy is uninterpretable without a selectivity check against a
memorization-based control task. In our work, RQ1 (decodability) and RQ3
(steering selectivity) directly adapt these two principles, applying
them to scoring rubrics rather than standard linguistic properties.

\section{Setup}\label{setup}

\textbf{Models.} We evaluate 25 open-weight, base LLMs
(non-instruction-tuned) ranging from 0.13B to 10.31B parameters with 16
to 42 layers across 14 model families, including Llama, Gemma, Falcon3,
SmolLM2/3, Granite, Mistral, OLMo, TAIDE, Apertus, and others (see
Appendix C, Table 6 for the full catalog). All 25 models are base models
by design. Although operational LLM-based scorers, such as those in the
Duolingo English Test, are typically fine-tuned models
(\citeproc{ref-Madnani2026-ScoringWritingSpeaking}{Madnani et al.,
2026}), we intentionally use base models to prevent instruction or
format tuning from confounding the rubric representations and score
output. This design establishes a clean baseline, enabling direct
comparisons with fine-tuned variants in future work.

The model parameter range also reflects a key deployment constraint
highlighted in the language-testing literature. Local institutions
evaluating AI-assisted assessment require practical model sizes for
on-premises hosting so that private response data remain local, rather
than being sent to massive cloud-hosted systems
(\citeproc{ref-Chuang2025-LanguageAssessmentEra}{Chuang and Yan, 2025}).

\smallskip

\textbf{BESTEP and its Scoring Rubrics.} This study uses the official
BESTEP writing rubrics, which human raters apply when scoring the test's
writing responses. BESTEP, which stands for ``the BEST test of English
Proficiency,'' was developed and is operated by the Language Training \&
Testing Center (LTTC) in support of Taiwan's national ``Program on
Bilingual Education for Students in College (BEST).'' The test assesses
the academic English proficiency of college students across four skills:
reading, listening, writing and speaking. We adopt the writing scoring
rubrics from the official test guidelines and have verified with LTTC
stakeholders that human raters apply these same public rubrics. The only
difference is that human raters complete training workshops with
authentic test responses before live scoring.

The writing section comprises three task parts, designed to evaluate
three distinct subconstructs. Part 1 (Answering Questions) is a short
factual-response task. Part 2 (Expressing Opinions) asks test-takers to
write emails expressing their opinions on campus events to stakeholders.
Part 3 (Integrated Writing) is the longest and most demanding task,
requiring test-takers to write a full essay.

Writing responses are scored with six analytic scoring rubrics, though
not every rubric applies to every part (see Table 1). Grammar and
vocabulary apply across all three parts, spelling and punctuation are
scored exclusively in the short-answer task (Part 1), and argument
relevance is evaluated solely in the integrated essay (Part 3).

\begin{table}[h]
\centering
\begin{tabular}{lccc}
\toprule
Scoring rubric & Part 1 & Part 2 & Part 3 \\
\midrule
Grammar & v & v & v \\
Vocabulary & v & v & v \\
Cohesion &  & v & v \\
Spelling/Punct. & v &  &  \\
Task Relevance & v & v &  \\
Argument Relevance &  &  & v \\
\bottomrule
\end{tabular}
\caption{Which scoring rubrics are scored in which task part.}
\end{table}

\smallskip

\textbf{Dataset.} All 25 models are probed and scored on the same fixed
set of 75 human-rated BESTEP writing responses (the gold-75), spanning
two test forms and three task parts (see Table 2).

\begin{table}[h]
\centering
\begin{tabular}{lrrrr}
\toprule
Form & Part 1 & Part 2 & Part 3 & Total \\
\midrule
Form 1 & 11 & 12 & 15 & 38 \\
Form 2 & 10 & 12 & 15 & 37 \\
\textbf{Total} & \textbf{21} & \textbf{24} & \textbf{30} & \textbf{75} \\
\bottomrule
\end{tabular}
\caption{Gold-75 essay coverage by test form and task part.}
\end{table}

\smallskip

\textbf{Activation extraction.} For every model, we extract
residual-stream activations at every layer for each of the gold-75
responses, taken at the final token of the grading prompt, where the
model would write its score. These activations are the shared input to
the probing (RQ1), contrastive-direction (RQ2), and steering (RQ3)
procedures. These procedures are described in the next three sections.

\smallskip

\textbf{Code and Data Availability.} Code, the analysis scripts, and the
result dataset behind every reported number are available at
\url{https://github.com/lopentu/inside-the-scorer-paclic2026}. They are
released for noncommercial research use under the PolyForm Noncommercial
License 1.0.0. However, the original and degraded essay pairs (Appendix
A) containing real exam responses are not yet public due to copyright
constraints. They are available for research purposes upon request.

\section{Experiments}\label{experiments}

\subsection{Score-Information
Representation}\label{score-information-representation}

\leavevmode\hspace*{\parindent}\textbf{RQ1.} Is the human-graded
holistic score decodable from a model's internal activations, and where
does that signal live? Which layers contain the highest-quality human
scoring signals?

\begin{figure}[h]
\centering
\includegraphics[width=\linewidth]{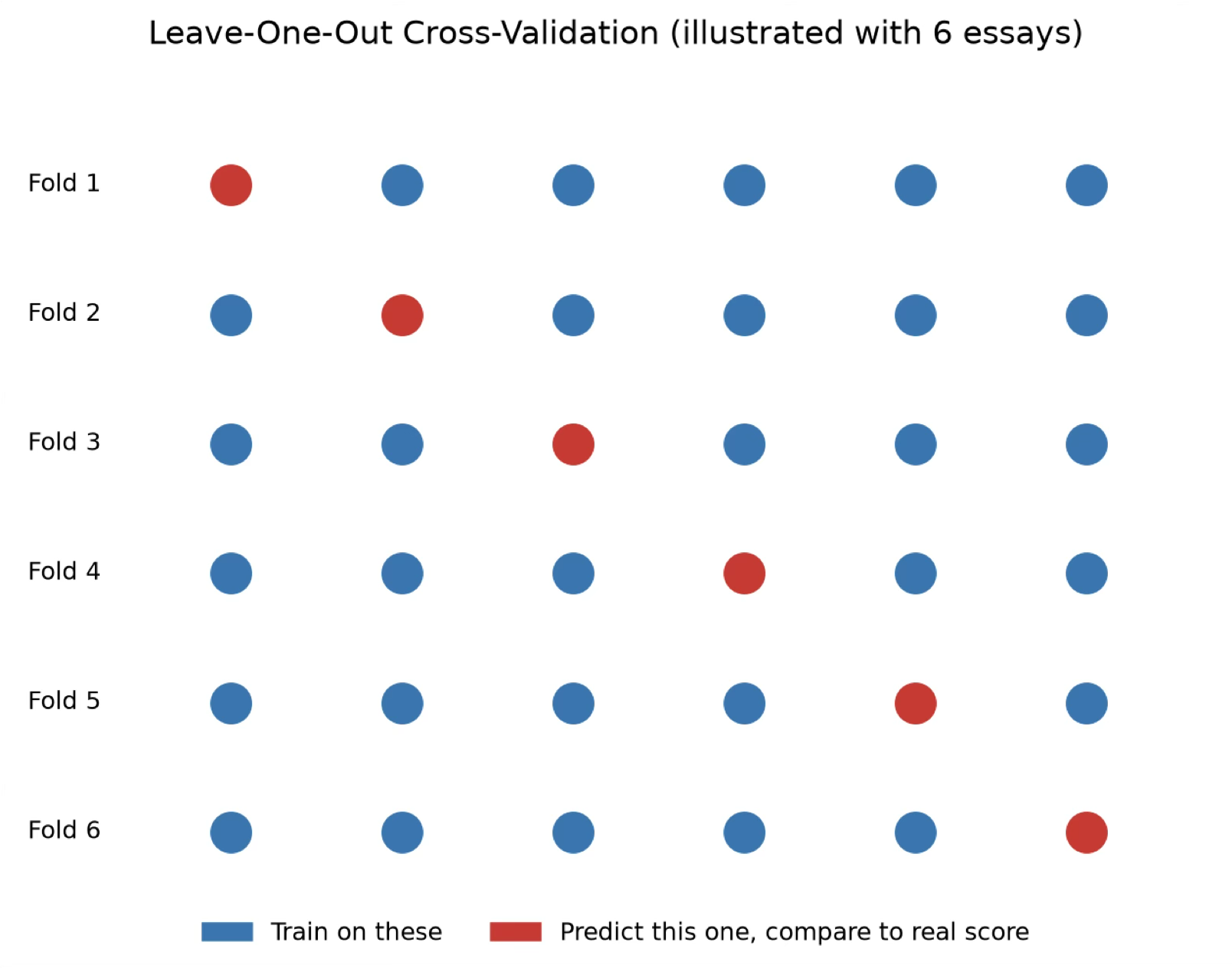}
\caption{Leave-one-out cross-validation (LOOCV).}
\end{figure}

\smallskip

\textbf{Method.} For each combination of {[}model, layer, part{]},
hereafter referred to as a cell, we train a Ridge regression probe on
final-token residual-stream activations to predict the human-rated
holistic band score. Each cell probe is evaluated using leave-one-out
cross-validation (LOOCV; Figure 1): in each fold, one response is held
out, and the cell probe is trained on the remaining \emph{N}-1 responses
(features centered and the ridge penalty selected within each training
round). All held-out predictions are pooled to compute a single
\emph{R}\textsuperscript{2} per cell (\emph{R}\textsuperscript{2} is
negative when predictions are worse than guessing the mean). We report
the peak-\emph{R}\textsuperscript{2} layer, the peak
\emph{R}\textsuperscript{2} itself, and the peak-to-final-layer drop,
treating the pooled-form scope (both test forms combined) as our primary
result. To verify that peak \emph{R}\textsuperscript{2} scores exceed
chance-level baselines from the probing procedure, each cell undergoes a
label-shuffling permutation test (1,000 score permutations). Because the
peak layer is selected as the maximum across 16 to 42 candidate layers,
we report the peak layer's \emph{p}-value both raw and
Bonferroni-corrected over that model's total layers.

\begin{figure}[h]
\centering
\includegraphics[width=\linewidth]{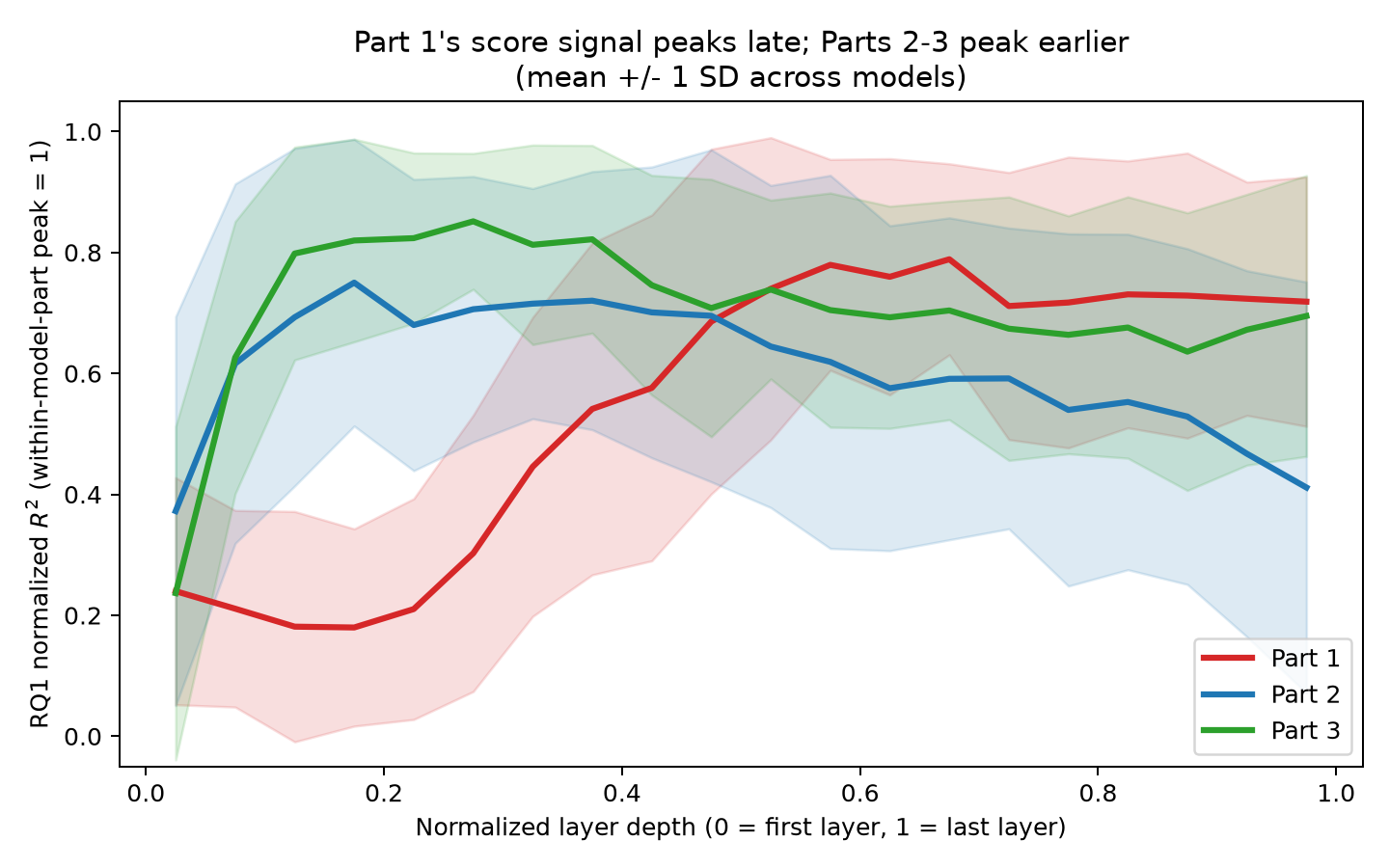}
\caption{Line chart of mean decodability (\emph{R}\textsuperscript{2}) by
normalized layer depth, one line per essay part, showing Part 1 peaking
latest.}
\end{figure}

\smallskip

\textbf{Results.} Human-rated score information is decodable in most
models, but not equally for every part, and not in a simple ``easy parts
early, hard parts late'' order. Against the permutation null, the peak
layer beats chance in all 25 models for Part 3 and in 22 of 25 for Part
2 after the layer correction, but in only 9 of 25 for Part 1 (24 of 25
uncorrected). We call a peak significant at \emph{p} \textless{} 0.05.
Raw \emph{p}-values sit at the permutation floor (\emph{p} = 0.001) for
25 of 25 models in Part 3, 21 in Part 2 and 9 in Part 1, so the
Bonferroni-corrected \emph{p}-values (medians 0.032, 0.034 and 0.096)
are limited by the 1,000 shuffles, not by the data. The one consistent
timing pattern: Part 1 (the shortest, short-answer task) peaks latest or
second-latest among the three parts in 23 of 25 models (Figure 2). This
holds in 8 of the 9 models whose Part 1 peak passes the corrected test,
but where it does not pass, the peak location is poorly estimated.
Across the 75 model-part cells, the signal usually fades toward the
final layer. The last layer keeps about two-thirds of the peak
\emph{R}\textsuperscript{2} (median drop 34\%), and in 14 out of the 75
cells it is worse than guessing the average score
(\emph{R}\textsuperscript{2} below zero).

Peak timing and decodability strength diverge across models. Part 1
produces the lowest peak \emph{R}\textsuperscript{2} in 24 of 25 models
and is never the highest among the three parts. Mean peak
\emph{R}\textsuperscript{2} across models scales with task length: 0.432
for Part 1, 0.633 for Part 2, and 0.773 for Part 3 (Table 3, Figure 3).
Part 1 is also the least stable across models
(peak-\emph{R}\textsuperscript{2} variance of 0.0338, versus 0.0095 for
Part 2 and 0.0038 for Part 3). This lower and more volatile human-score
decodability stems from Part 1 having the fewest gold responses
(\emph{n}=21), the shortest response lengths (about 47 words, vs.~about
117 for Part 2 and about 214 for Part 3), and a coarser rating scale,
yielding less textual evidence and fewer data points for probe
estimation.

\begin{table*}[!htb]
\centering
\begin{tabular}{lrrrrr}
\toprule
Part & Mean peak \emph{R}\textsuperscript{2} & Median peak \emph{R}\textsuperscript{2} & Peak-\emph{R}\textsuperscript{2} variance & Min & Max \\
\midrule
Part 1 & 0.432 & 0.485 & 0.0338 & 0.007 & 0.702 \\
Part 2 & 0.633 & 0.631 & 0.0095 & 0.413 & 0.877 \\
Part 3 & 0.773 & 0.773 & 0.0038 & 0.605 & 0.858 \\
\bottomrule
\end{tabular}
\caption{Peak \emph{R}\textsuperscript{2} statistics by task part, across 25
models.}
\end{table*}

\begin{figure}[h]
\centering
\includegraphics[width=\linewidth]{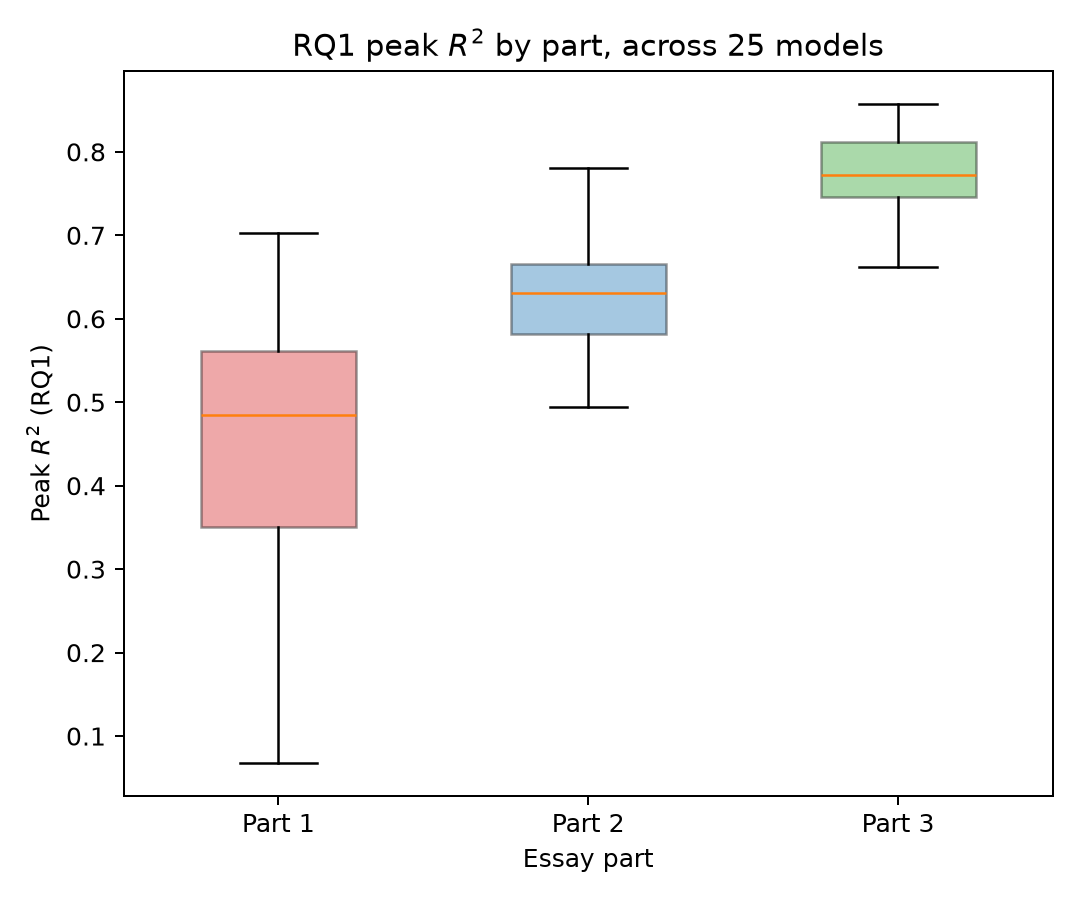}
\caption{Boxplot of peak \emph{R}\textsuperscript{2} across 25 models, grouped by
essay part, showing Part 3 highest and most consistent.}
\end{figure}

For the other two longer tasks, models carry information about essay
quality that a label-shuffle null cannot explain. For BESTEP's shortest
task type the signal peaks later, is weaker and less stable, and is
above the corrected null in only 9 of 25 models. We cannot explain this
pattern mechanistically here, but RQ2 returns to it. The Bonferroni
correction is conservative because adjacent layers are correlated, and
1,000 shuffles bound how small a \emph{p}-value can get (Limitations).

\subsection{Rubric Representation}\label{rubric-representation}

\leavevmode\hspace*{\parindent}\textbf{RQ2.} Do the six scoring rubrics
occupy distinguishable directions in representation space? If so, how
separated are they from each other?

\smallskip

\textbf{Method.} To evaluate rubric separability, RQ2 uses
rubric-targeted clean-versus-degraded essay mutations of the
gold-standard responses (\emph{n}=26). This is because holistic scores
blend individual rubrics and cannot isolate rubric-geometry dynamics on
their own. The mutations were generated by prompting Llama 3.3 (70B, via
Ollama) with one of 12 rubric-quoting specifications corresponding to
each valid {[}part, scoring rubric{]} combination (Table 1).

The source essays are the 26 gold responses rated at the top band of
Part 1's 1 to 5 scale (\emph{n}=5) and at a high band on the 1 to 6
scales of Part 2 (\emph{n}=9) and Part 3 (\emph{n}=12). Each is degraded
once for every rubric scored in its part. Each specification quotes the
official BESTEP band descriptors for a target rubric and instructs the
model to degrade the response to a low band along that single dimension
while holding all other dimensions, overall meaning, and response length
(within 10\%) constant. Each mutated response was verified by two
trained BESTEP human raters independently against its target dimension
prior to analysis (see Appendix A for full procedures and worked
examples).

For each {[}model, layer, part, rubric{]} cell, we estimate a
contrastive direction: the mean activation of the clean essays minus the
mean activation of their degraded versions, taken at the same position
as in RQ1 and scaled to unit length. Each direction rests on 5, 9, or 12
essay pairs (Parts 1 to 3). To measure how stable a direction is, we
split its pairs at random into two halves, estimate a direction from
each half, and take the cosine similarity between the two; the average
over 20 random splits is the split-half reliability. Two directions are
compared by their cosine similarity at the same layer, and a comparison
is kept only if both directions have a split-half reliability of at
least 0.6. For each direction, the geometry gap is its mean similarity
to the same rubric in the other parts (same rubric) minus its mean
similarity to the other rubrics in the same part (different rubric).
Different rubrics are compared only within a part, so that a difference
between two rubrics is not confused with a difference between tasks or
essays. A positive gap means that a rubric's direction is more similar
to itself than to its neighbors, that is, that it has a direction of its
own.

\smallskip

\textbf{Results.} Rubric directions are separate in the two longer tasks
(Parts 2 and 3) but not in Part 1. A positive geometry gap means that a
rubric's direction is more similar to itself than to the other rubrics
in the same part. Across 1,470 reliability-gated layer-part cells, the
mean gap is 0.334, and 74.5\% of cells are positive. This average covers
only rubrics scored in more than one part; spelling/punctuation and
argument relevance are scored in a single part, have no same-rubric
comparison across parts, and are not included. The gap differs sharply
by part. In Part 2 (mean gap 0.372, 89.7\% of cells positive) and Part 3
(0.609, 99.2\%), the rubrics have directions of their own. In Part 1 the
mean gap is negative (−0.220, 6.2\% of cells positive), especially for
grammar and vocabulary. This suggests that different rubrics are more
alike than a single rubric is across parts.

Pair by pair, Table 4 gives the similarity between each two rubrics'
directions: 1 means the same direction, 0 means unrelated, and a
negative value means partly opposite directions. Because each direction
rests on only 5 to 12 essay pairs (made from high-band gold essays) and
carries noise, we correct the similarities for it (disattenuated cosine
similarity). Among the 15 possible rubric pairs, 12 can be compared. The
other three (argument relevance with spelling/punctuation, argument
relevance with task relevance, and cohesion with spelling/punctuation)
are never simultaneously used to score in the same part. The 12
comparable pairs range from −0.102 to 1.0. The most separate pair is
\texttt{argument\_relevance} with \texttt{grammar} (−0.102,
anti-correlated beyond measurement noise), and
\texttt{argument\_relevance}, a Part-3-only rubric, appears in three of
the five least similar pairs. The most alike are \texttt{grammar} with
\texttt{spelling\_punctuation} (1.0, statistically indistinguishable
from the same rubric) and \texttt{spelling\_punctuation} with
\texttt{vocabulary} (0.984). Both are Part-1-only comparisons, in line
with Part 1's negative gap.

In short, rubric separation is clear in Parts 2 and 3, most visibly pair
by pair for \texttt{argument\_relevance}, and absent in Part 1.

\begin{table}[h]
\centering
\footnotesize
\begin{tabular}{lrrl}
\toprule
Rubric pair & \emph{n} & Sim. & 95\% CI \\
\midrule
Arg. Rel. x Gram. & 564 & −0.10 & {[}−0.13, −0.08{]} \\
Spell. x Task Rel. & 6 & 0.01 & {[}−0.36, 0.37{]} \\
Arg. Rel. x Cohes. & 586 & 0.08 & {[}0.06, 0.11{]} \\
Gram. x Task Rel. & 416 & 0.28 & {[}0.25, 0.32{]} \\
Arg. Rel. x Vocab. & 419 & 0.31 & {[}0.29, 0.34{]} \\
Cohes. x Gram. & 800 & 0.39 & {[}0.35, 0.42{]} \\
Gram. x Vocab. & 823 & 0.49 & {[}0.46, 0.52{]} \\
Task Rel. x Vocab. & 336 & 0.51 & {[}0.48, 0.55{]} \\
Cohes. x Task Rel. & 334 & 0.52 & {[}0.48, 0.55{]} \\
Cohes. x Vocab. & 691 & 0.61 & {[}0.58, 0.65{]} \\
Spell. x Vocab. & 192 & 0.98 & {[}0.96, 1.00{]} \\
Gram. x Spell. & 328 & 1.00 & {[}1.00, 1.00{]} \\
\bottomrule
\end{tabular}
\caption{Disattenuated rubric-pair similarity (1.0 = same rubric, 0 = unrelated),
sorted low to high. Gram. = grammar, Vocab. = vocabulary, Cohes. =
cohesion, Spell. = spelling/punctuation, Task Rel. = task relevance,
Arg. Rel. = argument relevance.}
\end{table}

\FloatBarrier

\subsection{Causal Specificity}\label{causal-specificity}

\leavevmode\hspace*{\parindent}\textbf{RQ3.} Does steering a rubric's
direction move the model's score specifically for that scoring rubric,
or does any rubric-adjacent push move it about equally?

\smallskip

\textbf{Method.} RQ3 tests the rubric directions from RQ2 causally, not
only representationally. We add a direction to the residual stream at
one layer, at every token position of the grading prompt. The direction
has unit length and is scaled by alpha times the mean L2 norm of that
layer's residual stream, measured at the answer position of the clean
and degraded essay prompts (alpha = 0.05, 0.1, 0.25, 0.5), so the
strength is comparable across layers and models. The outcome is the
change in the model's own expected band score. At the answer position of
the grading prompt, where the model would write its score, we take the
model's logits for the digit tokens 0 to the maximum band (5 for Part 1,
6 for Parts 2 and 3), convert them to probabilities with a softmax over
those digits only, and compute the probability-weighted mean of the
digit values. This takes one forward pass and uses neither a separate
scoring prompt nor a learned readout. The outcome is this expected band
with the direction injected minus the same quantity without it (the
unsteered baseline). For degraded essays with a known target rubric, we
inject four kinds of direction:

\begin{itemize}
\tightlist
\item
  the matching rubric direction (self);
\item
  a mismatched rubric direction (other);
\item
  an orthogonal control, a rubric direction with the axis shared by all
  rubrics in the part removed, which tests whether overlap between
  directions hides specificity;
\item
  a generic holistic direction, the direction the RQ1 probe uses to
  predict the human band score, which tests whether any score-raising
  push works.
\end{itemize}

Self and other effects are paired at the finest valid unit, {[}model,
layer, part, form, degradation, alpha{]}, which gives 74,304 paired
cells across the 25 models. We test whether the matching direction beats
a mismatched one by a practically meaningful margin, which we define as
more than 0.05 score points or more than 10\% of the self effect.

\smallskip

\textbf{Results.} Steering moves the score for five of the six rubrics,
but not selectively: the direction that matches the essay's degraded
rubric works barely better than a mismatched one.

\emph{Steering works.} Grammar, vocabulary, cohesion,
spelling/punctuation, and task relevance each have a clearly positive
effect when their own direction is injected (self effect, 0.168 to 0.359
score points). Only \texttt{argument\_relevance} has no detectable self
effect (−0.083, 95\% CI {[}−0.195, 0.024{]}): its direction neither
raises nor lowers the score.

\emph{Steering is not selective.} Pooled across all paired cells, the
matching direction beats a mismatched one by only 0.0074 score points
(95\% CI {[}0.0036, 0.0117{]}), which is 1.74\% of the absolute self
effect (95\% CI {[}0.85\%, 2.74\%{]}). Both confidence intervals lie
well inside our margin, so the advantage is statistically detectable but
practically negligible. This holds under every robustness check we ran:
at the cell level (0.0085, 95\% CI {[}0.0051, 0.0119{]}); when any one
model or any one model family is left out (never above 0.0083); and in a
cluster bootstrap that resamples whole model families to account for
shared training lineage, such as the three Llama-3.x checkpoints
(0.0073, 95\% CI {[}0.0035, 0.0108{]}).

\emph{Pair by pair.} The pooled number combines two patterns (Figure 4).
First, among the four rubrics that both steer the score and are reliably
measured (grammar, vocabulary, cohesion, and spelling/punctuation), the
matching direction shows no detectable advantage over a mismatched one
in any of the 10 testable head-to-head comparisons. The other two
ordered pairs, cohesion with spelling/punctuation, cannot be tested
because the two are never degraded in the same part. In one of the ten,
vocabulary detectably loses to spelling/punctuation, not because
vocabulary is weak, but because spelling/punctuation is the strongest
general-purpose lever of the six. Second, \texttt{argument\_relevance}
loses all three of its comparisons, because its own direction never
moves the score.

\emph{Within each task part.} We also compared every pair of rubrics
scored in the same part (18 pairs in all). For each pair, we added the
two directions' gaps, one on each rubric's own essays. This cancels a
rubric's general strength as a lever and leaves only the advantage of
matching. The matching advantage ranges from −0.015 to 0.030 score
points. It is detectable in 9 of the 18 pairs and reversed in one, but
it is small: in the detectable pairs it is 2\% to 14\% of the pair's own
effects (median 6\%), and the largest values involve the weak Part 1
task-relevance direction.

\begin{figure*}[!htb]
\centering
\includegraphics[width=0.70\linewidth]{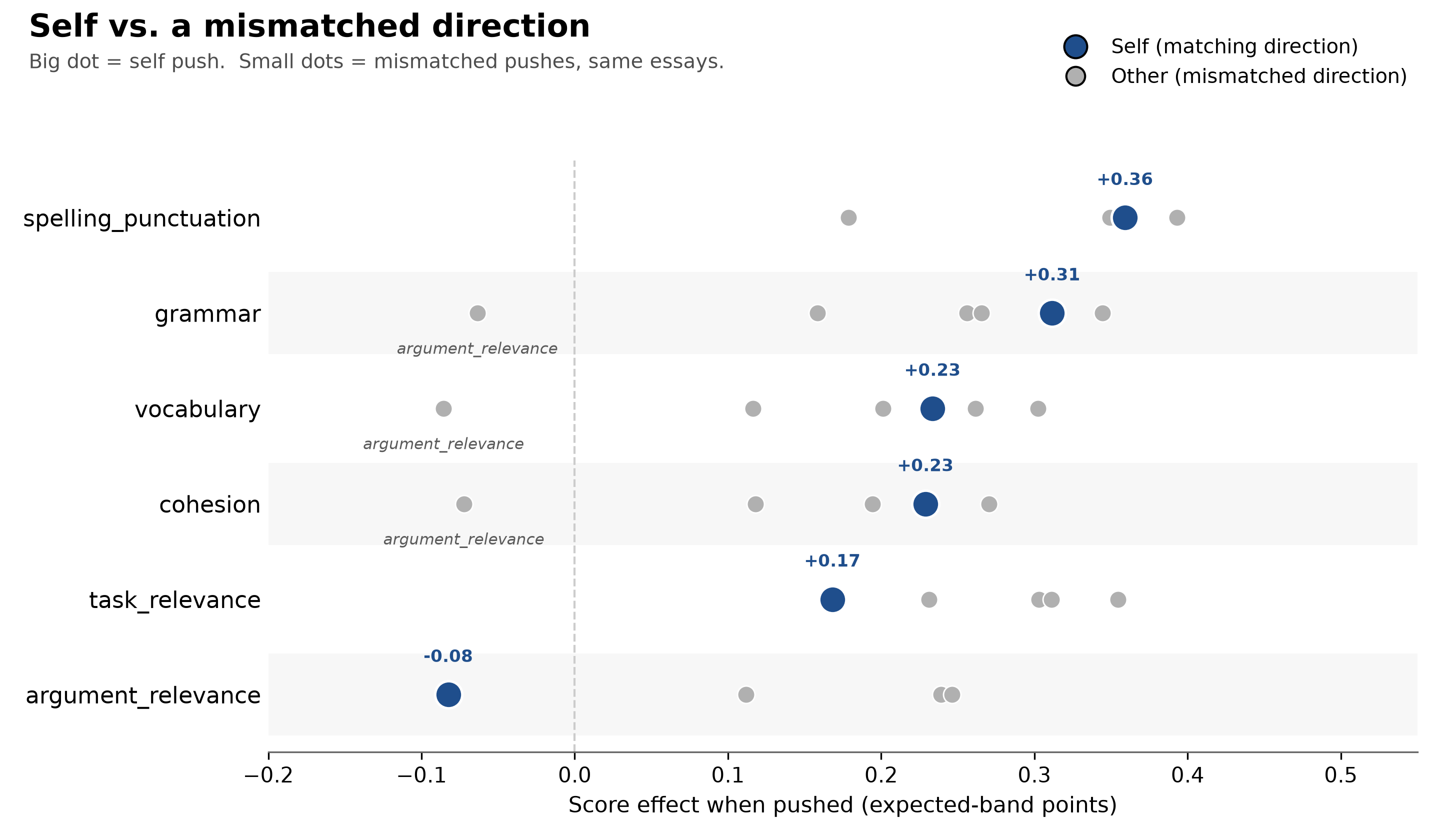}
\caption{Scatter/dot plot per scoring rubric comparing the self
(matching-direction) steering effect against the distribution of
mismatched-direction effects, showing that the self effects of grammar,
vocabulary, cohesion, and spelling/punctuation overlap the mismatched
effects, while task relevance is lower and argument relevance is
negative.}
\end{figure*}

\emph{Controls.} Two controls calibrate these effects (Table 5), where
Delta is the mean change in the expected band score in score points and
Slope is how much that change grows per unit of alpha, both averaged
over models. The orthogonal control is not a flat baseline: it lowers
the score slightly (Delta −0.097) and more so at larger alpha (Slope
−0.288). So the model is not indifferent to an injected direction, only
to which rubric it targets, and the score-raising effect lies mostly in
the axis that the rubric directions share (Check 2 below). The generic
holistic direction raises the score (Delta 0.130) but less than the self
or other direction at every alpha tested, which argues against a pure
``generic tampering'' account of the specificity gap above. Its alpha
slope (0.550) is close to those of self (0.589) and other (0.562), so it
behaves like a weaker member of the same family of levers. At the
strongest alpha (0.5), the average self effect is 0.361 score points,
about 6\% of the 6-point band scale, a moderate intervention.

\begin{table}[h]
\centering
\footnotesize
\begin{tabular}{lrr}
\toprule
Metric & Orthogonal & Band probe \\
\midrule
Delta & −0.097 & 0.130 \\
Delta 95\% CI & {[}−0.192, −0.020{]} & {[}0.024, 0.225{]} \\
Slope & −0.288 & 0.550 \\
Slope 95\% CI & {[}−0.519, −0.063{]} & {[}0.251, 0.846{]} \\
\bottomrule
\end{tabular}
\caption{Pooled delta and alpha-sensitivity for the orthogonal-control and
generic holistic-direction conditions, same 25-model cohort and pairing
as the self/other results.}
\end{table}

\emph{Link to RQ2.} A pair's separation in RQ2 (Table 4) predicts how
large the causal difference for that pair is under steering
(\emph{r}=0.717 across the 24 ordered pairs, that is, the 12 pairs in
both directions), but it does not predict which direction wins. The
evidence for separation and for nonselective steering also comes partly
from different rubrics: \texttt{argument\_relevance} is among the most
separate rubrics in RQ2 but has no steering effect, whereas the four
rubrics that steer and are reliably measured have moderately to highly
similar directions (Table 4), so the two results should be read together
(Discussion).

\emph{Could noise or overlap hide a larger specificity?} RQ2's
directions are reliability-gated, but RQ3 injects the raw,
non-disattenuated directions regardless of noise, and different rubrics'
directions are already substantially correlated (Table 4). Noise and
this overlap could mask a larger true specificity, so we ran three
checks.

\emph{Check 1: more reliable directions.} The specificity ratio is 1.8\%
across all cells (a cell-level figure, which is why it differs slightly
from the model-level 1.74\% above), 5.4\% for cells that meet the RQ2
reliability gate (at least 0.6), 3.1\% at a reliability of at least 0.7,
and 8.3\% in the top reliability quartile (about 0.85 or higher; self
0.670 vs.~other 0.605, gap 95\% CI {[}0.057, 0.072{]}). The rise is not
monotone, but the most reliable directions show the highest ratio, so
less reliable directions may understate true specificity. Even the
best-measured quarter shows self only modestly outperforming other.

\emph{Check 2: remove the shared axis.} We remove the axis shared by all
rubrics in a part from each direction (the orthogonal-control condition)
and repeat the self-versus-other comparison on these overlap-corrected
directions, using data already collected. If overlap were hiding real
independent control, this would reveal it, and it does not. Both the
self and other effects are slightly negative (−0.093 and −0.100), and
their gap (0.0064, 95\% CI {[}0.0003, 0.0125{]}) is about 5\% of the
mean absolute size of the corrected self effect (0.133 across models,
larger than the −0.093 average because the sign differs across models),
close to the 5.4\% in Check 1. Once the shared component is removed, the
rubric-specific residual is not a strong positive lever for either
rubric, so most of the raw effect was the shared component, not overlap
hiding independent control.

\emph{Check 3: overlap with one specific rival.} For each pair of
rubrics, we strip from each direction its projection onto the other, so
that the matching and mismatched directions lose exactly their mutual
overlap (pairs too collinear to leave a residual are skipped). We inject
the two residuals next to a norm-matched random direction as a noise
floor. Across all 25 models (884 pair cells), both residuals raise the
score above the noise floor, by 0.194 (matching) and 0.181 (mismatched)
points, so rubric-specific residual structure exists and moves the
score. This does not conflict with Check 2. The random-direction noise
floor is itself negative (−0.101), so the corrected directions in Check
2 (−0.093) sit at the noise level, while the residuals here (+0.093 and
+0.080) sit well above it, probably because removing only one rival's
overlap leaves part of the lever that all rubrics share. The matching
residual still does not win: the gap is 0.013 points (95\% CI {[}−0.015,
0.042{]}), about 4.5\% of the average absolute matching effect (about
0.30, since the 0.194 above is measured from a noise floor that is
itself negative), and the interval includes zero. Pairs with a cosine
below and at or above 0.6 give the same picture (gaps of 0.012 and
0.014). Removing the overlap pair by pair leaves the levers
interchangeable.

\emph{Models that agree best with human raters.} These base models agree
only modestly with the human band (Limitations), so shared control could
reflect weak scorers. Among the seven models whose own scores correlate
with the human band at \emph{r} ≥ 0.6 (an exploratory cut-off set after
seeing the agreement values), the matching direction beats a mismatched
one by 0.011 score points (95\% CI {[}0.007, 0.015{]}), which is 1.9\%
of the absolute self effect, against 1.74\% in the full set. Models with
better agreement therefore show the same shared control.

\FloatBarrier

\section{Discussion}\label{discussion}

Taken together, the three research questions give one account of what a
rubric-based LLM scorer does inside. The human score can be read out of
the models' activations (RQ1). This is clear for the two longer tasks,
where the peak layer beats a label-shuffle null after correction in 25
of 25 models (Part 3) and 22 of 25 (Part 2), and only tentative for the
shortest task, where 9 of 25 models do. The models also represent at
least some scoring rubrics as partly separable directions, rather than
as one undifferentiated quality signal (RQ2). But separable
representations do not give independent control (RQ3). Steering a
rubric's direction reliably moves the score, yet for the rubrics that
are both effective and reliably measured, pushing the matching direction
is essentially interchangeable with pushing a different one. The model
appears to exercise one shared control with five working handles of
different strength, not six independent, rubric-specific ones.

This matters for how neural AES is justified. Operational providers that
need control and verifiability get them by design. In a decomposed
system, each subconstruct has its own slot, and a neural feature can
influence the score only through that slot
(\citeproc{ref-Madnani2026-ScoringWritingSpeaking}{Madnani et al.,
2026}). An end-to-end LLM scorer has no such slot, so our audit asks
whether it contains something equivalent on its own. In these base
models it does not. Rubric information is present and partly separable,
but it is not exposed as separate controls. High agreement with human
raters, or the mere decodability of rubrics, is therefore not evidence
that a model uses those rubrics as separately controllable levers.
Controllability is a diagnostic that score agreement cannot supply, and
the models that agree best with human raters, discussed below, show this
directly.

These results also bear on the rubrics themselves. Analytic rubrics are
written for human raters, who are trained to treat each criterion as a
separate judgment. In these base models, the criteria are partly
represented, with separable directions for the two longer tasks, but
they do not work as separate levers. None of the 10 testable comparisons
among the four reliably measured rubrics shows a detectable advantage
for the matching direction, and a within-part test finds only small
advantages in some pairs (typically 6\% of the pair's own effects, at
most 14\%; RQ3). In that sense, rubrics designed for human raters do not
behave as independent controls in base LLMs.

Why the control is shared is an open question, because there are two
readings. In one, the model collapses rubrics that the test keeps apart.
In the other, the model faithfully reproduces the structure of human
analytic ratings, in which dimensions are known to move together. Halo,
the carry-over of a rater's overall impression into separate trait
scores, is a documented rater effect in the scoring of written
composition
(\citeproc{ref-EngelhardJr1994-ExaminingRaterErrors}{Engelhard Jr,
1994}; \citeproc{ref-Myford2004-DetectingMeasuringRater}{Myford and
Wolfe, 2004}), and its relation to rating accuracy is itself contested
(\citeproc{ref-Fisicaro1988-ReexaminationRelationHalo}{Fisicaro, 1988}).
Our gold set carries only holistic band scores, so we cannot separate
the two readings here. Either way, a provider cannot assume that rubric
labels from a neural scorer are separately controllable without
checking.

One might worry that shared control simply reflects weak scorers, since
these base models agree only modestly with human raters (Limitations).
The data do not suggest that. Among the seven models whose own scores
correlate with human scores at \emph{r} of 0.6 or higher, the matching
direction still beats a mismatched one by only 0.011 score points (95\%
CI {[}0.007, 0.015{]}; 1.9\% of the absolute self effect), about the
same as in the full set (RQ3). Higher agreement with human raters did
not bring independent control.

The evidence for separation and for nonselective steering also comes
from partly different rubrics, so the two should be read together.
\texttt{argument\_relevance} appears in three of the five least similar
pairs in RQ2, but its steering effect is undetectable. The four rubrics
that steer and are reliably measured, in contrast, have moderately to
highly similar directions (Table 4). Nonselective steering among
overlapping directions is what overlap alone would predict, which is why
the pairwise check in RQ3 matters. With each pair's overlap removed,
both residuals still move the score and still do so about equally.

A related result concerns what the score responds to. The generic
holistic direction, learned from the human band scores, raises the
model's score less than the rubric directions do at every strength we
tested (Table 5), and the strongest lever of all is
spelling/punctuation, a surface feature. This suggests, but does not
show, that the model's own score leans more on surface language features
than on the quality signal that RQ1 can read out of its activations. The
decodable signal also usually fades toward the final layer, with a
median drop of 34\% from the peak and worse than guessing the average
score in 14 of 75 model-part cells. Information that is decodable is not
necessarily information that the model uses.

Finally, the pattern is not uniform across the assessment. Part 1,
BESTEP's shortest task, shows the weakest and least stable score signal
(RQ1; only 9 of 25 models pass the corrected permutation test, so its
later peak is tentative) and no meaningful rubric separation (RQ2). The
two longer, more integrated parts support a stronger signal and real, if
not independently controllable, rubric structure. Validity claims should
be checked at that grain, not assumed to generalize across task format.

\section{Limitations}\label{limitations}

All 25 models are base (non-instruction-tuned) models. We chose them as
a baseline, but this means we cannot say how instruction-tuned, chat, or
task-fine-tuned variants, which resemble deployed scorers more closely,
would behave. The gold set is 75 essays from one assessment (BESTEP),
which limits essay-level diversity, and model coverage tops out at
10.31B parameters, so the results may not hold for substantially larger
models. All models also see the same grading prompt, and we steer by
adding one direction at one layer, so other prompts or other kinds of
intervention could give different results.
\texttt{argument\_relevance}'s own steering direction never reaches
significance; whether that reflects a representational gap or a rubric
that needs more than surface language features to steer is open. More
models, other writing assessments, and non-English rubric-based
assessments could test whether the findings travel.

This paper tests how scoring rubrics are represented and controlled
inside the models, not whether the models score like human raters. The
human band score is a fixed reference only: it is the target of the RQ1
probe and the source of the generic holistic direction in RQ3. The
models' own scores agree only modestly with it (pooled Pearson \emph{r}
across the 75 essays: median 0.41, 7 of 25 models at \emph{r} ≥ 0.6,
none at 0.8), so the findings describe imperfect zero-shot base scorers,
not validated scoring systems. Our check on the seven best-agreeing
models is also exploratory. It is small, and we chose the cut-off of 0.6
after seeing the agreement values.

Two limits concern the degraded essays that define every rubric
direction. The first is their number: all 104 come from only 26 gold
responses at band 5, so each direction rests on 5 to 12 essay pairs. The
second is how they were made. One LLM generated them, and two trained
BESTEP graders then checked every one of the 104. Each grader,
separately, saw an original and its degraded version, was told the essay
had been degraded on one scoring rubric, and had to name which, and both
confirmed the intended rubric every time (Appendix A). That shows the
target rubric is recognizable, not that nothing else changed. The
grammar/spelling\_punctuation similarity ceiling in Table 4, and the
near-ceiling spelling/vocabulary pair, can arise because errors in one
dimension leave traces in another even in ``pure'' stimuli (Appendix A
gives an example), so we do not read those pairs as evidence about the
model alone. Cleaner single-dimension manipulations are future work.

Because the gold set carries no rubric-level human ratings, we cannot
test whether shared control reflects the models or the halo structure of
human analytic ratings (Discussion). A disattenuated human inter-rubric
correlation matrix would let a future study separate the two.

Several statistical limits apply. The RQ1 correction for layer selection
is Bonferroni, which is conservative because adjacent layers are
correlated, and 1,000 label shuffles set a floor on each \emph{p}-value;
a max-over-layers permutation test would be exact. The paired-cell
analysis is not a simple random sample, because layers, alpha values,
and models sharing a training lineage are correlated. The primary
inference averages within model before pooling, and the family-clustered
bootstrap gives an essentially unchanged estimate, but a full
hierarchical model across family, essay, rubric, and layer would be a
stronger test. The pairwise check removes overlap between two directions
at a time, and pairs too collinear to leave a residual are skipped, so
the most entangled pairs are the ones it cannot speak to. How best to
isolate one direction's rubric-specific component from a correlated one
is itself an active methodological question
(\citeproc{ref-Elazar2021-AmnesicProbingBehavioral}{Elazar et al.,
2021}). Finally, ``no detectable advantage'' is not proof of equivalence
for a single pair, because the intervals for single pairs are wide (for
example, {[}−0.018, 0.112{]} score points for grammar against
vocabulary); only the pooled estimate is tight. We also did not correct
for multiple comparisons across the 10 head-to-head and 18 within-part
comparisons, and the correlation of 0.717 in RQ3 counts each of the 12
pairs twice, so its effective sample is about 12 pairs.

\section{Conclusion}\label{conclusion}

LLM-based essay scoring is attractive but opaque, and high-stakes
testing asks for control and verifiability. We examined base LLMs
prompted as scorers from the inside, asking whether their scoring
rubrics are represented and whether they can be controlled
independently. Across 25 open-weight base models, holistic scores are
decodable (clearly for the two longer task types, weakly for the
shortest) and scoring rubrics are partly separable, but neither property
implies independent causal control. Steering a rubric's direction moves
the score as a general lever for all four rubrics that this design can
measure reliably (and for five of the six overall), and removing each
pair's overlap does not change that. The same holds among the models
whose scores agree best with human raters. This may reflect the models,
the halo structure of human ratings, or both, and rubric-level human
ratings would be needed to tell them apart. Any rubric-based LLM scorer
should therefore be examined for independent controllability before its
rubric labels are treated as separately meaningful, and on this
evidence, base LLMs do not yet offer it. Instruction-tuned and
fine-tuned scorers are the natural next comparison.

\appendix
\renewcommand{\thesection}{Appendix \Alph{section}}

\section{Degradation Examples}\label{degradation-examples}

RQ2's contrastive directions are estimated from clean-versus-degraded
essay pairs (§RQ2 Method). Each degradation is produced by prompting
llama3.3:70b with a spec that quotes the official BESTEP low-band
descriptor for one scoring rubric and instructs the model to hold every
other dimension constant. The first example below degrades one source
essay on two adjacent scoring rubrics, and the second shows a cohesion
degradation, to make the intended separation (and its limits) concrete.

\smallskip

\textbf{Part 1, grammar (target: verb form, agreement, word order
only).}

\begin{itemize}
\tightlist
\item
  \emph{Original}: ``Three tours will be given. What should I prepare
  for the training? The place in my school that I visit most often is
  the classroom, which is the place I learn knowledges about my major.''
\item
  \emph{Grammar-degraded}: ``Three tour will be give. What I should
  prepare for the training? The place in my school that I visiting most
  often are the classroom, which the place I learn knowledges about my
  major.''
\item
  \emph{Spelling/punctuation-degraded (same source sentence)}: ``Three
  tuors will be given. What shoud I prepair for the trainning. the place
  in my school that i visit most often is the clasroom which is the
  place i lern knowleges about my major.''
\end{itemize}

The grammar mutant changes verb forms, subject-verb agreement, and word
order while leaving spelling intact; the spelling/punctuation mutant
misspells content words and drops punctuation while leaving grammar
intact, but it also drops sentence-initial capitalization (``the
place'', ``i visit''), a surface feature a grammar-sensitive signal
could partly pick up on. This is a concrete illustration of the
limitation discussed below: even a mutation that targets one dimension
by instruction can leave a residual trace in an adjacent one.

\begin{figure*}[!htb]
\centering
\includegraphics[width=0.80\linewidth]{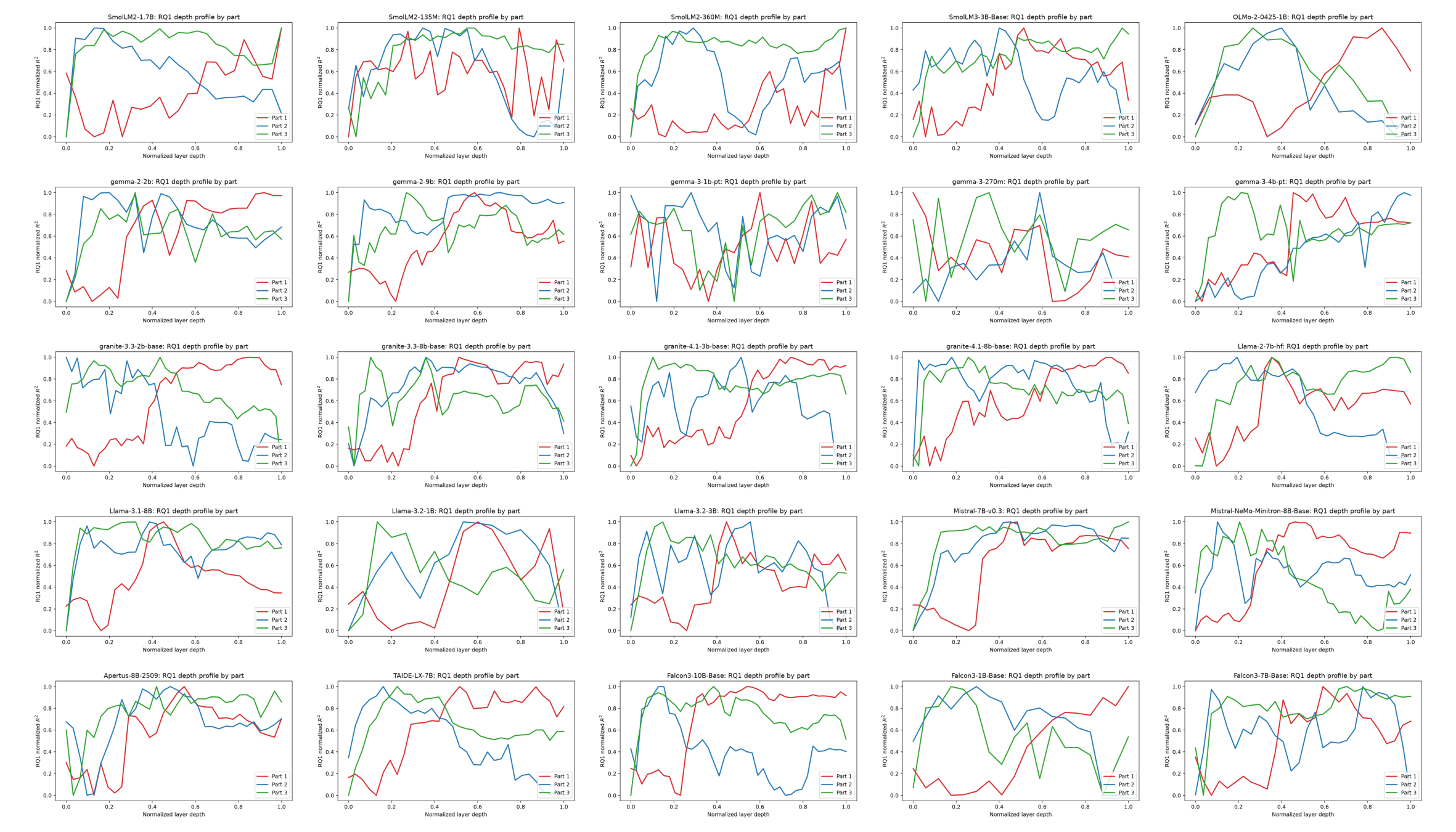}
\caption{Grid of 25 small-multiple depth-profile plots, one per model, for
per-model inspection of the pooled RQ1 pattern.}
\end{figure*}

\smallskip

\textbf{Part 2, cohesion (target: connectives and sentence-joining
only).}

\begin{itemize}
\tightlist
\item
  \emph{Original}: ``I wrote this email in regards to the new food court
  on campus. First of all, I want to say it is a good change for the
  school because the food court can help save the time, the students
  won't need to leave the campus to get food. {[}\ldots{]} In this way,
  the variety of the options and the opening hours might be appealing to
  more students.''
\item
  \emph{Cohesion-degraded}: ``I wrote this email about the new food
  court on campus. It is a good change for the school. The food court
  can help save the time. The students won't need to leave the campus to
  get food. {[}\ldots{]} The variety of the options might be appealing
  to more students. The opening hours might be appealing to more
  students.''
\end{itemize}

Here the degradation removes connectives (``First of all'', ``In this
way'') and splits joined sentences into flat, simple ones, with no
change to individual sentences' grammar or vocabulary, a cleaner
separation than the grammar/spelling pair above, consistent with
cohesion's lower same-rubric entanglement in Table 4.

\smallskip

\textbf{Verification.} Every degraded essay was checked by two trained
BESTEP essay graders before use. Working separately, each grader was
shown an original essay together with its degraded version, was told
that the degraded version had been downgraded on one scoring rubric, and
had to identify which. All 104 degraded essays (20 for Part 1, 36 for
Part 2, and 48 for Part 3, that is 5, 9, or 12 for each part and scoring
rubric, made from 26 band-5 gold responses) were checked this way. Both
graders confirmed the intended scoring rubric for every one, a consensus
of 100\%, and the check was run as a random double sample and as a
cross-comparison between the graders. This establishes that the intended
scoring rubric is recognizable in each degraded essay. It does not
establish that no other dimension changed, and the grammar/spelling
example above shows that adjacent dimensions can carry a trace. The
degraded essay pairs contain exam responses and are not yet public; they
are available from the authors for research purposes on request.

\section{Per-Model Depth Profiles}\label{per-model-depth-profiles}

Figure 5 shows the RQ1 depth-profile curve (Figure 2's pooled version,
disaggregated) for each of the 25 models individually, for readers who
want to check whether the pooled pattern in Figure 2 holds for a
specific model rather than only in aggregate.

\FloatBarrier

\section{Model Catalog}\label{model-catalog}

Table 6 lists the 25 open-weight base models probed and steered in this
study, sorted by parameter count.

\begin{table*}[!htb]
\centering
\begin{tabular}{llrrl}
\toprule
Model & Org & Params (B) & Layers & Published \\
\midrule
SmolLM2-135M & Hugging Face & 0.13 & 30 & 2024-10 \\
gemma-3-270m & Google & 0.27 & 18 & 2025-08 \\
SmolLM2-360M & Hugging Face & 0.36 & 32 & 2024-10 \\
gemma-3-1b-pt & Google & 1.00 & 26 & 2025-03 \\
Llama-3.2-1B & Meta & 1.24 & 16 & 2024-09 \\
OLMo-2-0425-1B & AI2 & 1.48 & 16 & 2025-04 \\
Falcon3-1B-Base & TII & 1.67 & 18 & 2024-12 \\
SmolLM2-1.7B & Hugging Face & 1.71 & 24 & 2024-10 \\
granite-3.3-2b-base & IBM & 2.53 & 40 & 2025-04 \\
gemma-2-2b & Google & 2.61 & 26 & 2024-07 \\
SmolLM3-3B-Base & Hugging Face & 3.08 & 36 & 2025-07 \\
Llama-3.2-3B & Meta & 3.21 & 28 & 2024-09 \\
granite-4.1-3b-base & IBM & 3.40 & 40 & 2026-04 \\
gemma-3-4b-pt & Google & 4.30 & 34 & 2025-03 \\
Llama-2-7b-hf & Meta & 6.74 & 32 & 2023-07 \\
TAIDE-LX-7B & TAIDE (NARLabs/NCHC) & 6.94 & 32 & 2024-05 \\
Mistral-7B-v0.3 & Mistral AI & 7.25 & 32 & 2024-05 \\
Falcon3-7B-Base & TII & 7.46 & 28 & 2024-12 \\
Apertus-8B-2509 & Swiss AI Initiative & 8.00 & 32 & 2025-09 \\
Llama-3.1-8B & Meta & 8.03 & 32 & 2024-07 \\
granite-3.3-8b-base & IBM & 8.17 & 40 & 2025-04 \\
granite-4.1-8b-base & IBM & 8.38 & 40 & 2026-04 \\
Mistral-NeMo-Minitron-8B-Base & NVIDIA & 8.41 & 40 & 2024-08 \\
gemma-2-9b & Google & 9.24 & 42 & 2024-06 \\
Falcon3-10B-Base & TII & 10.31 & 40 & 2024-12 \\
\bottomrule
\end{tabular}
\caption{The 25-model catalog (0.13B-10.31B parameters, 14 families).}
\end{table*}

\section*{References}\label{bibliography}
\addcontentsline{toc}{section}{References}

\protect\phantomsection\label{refs}


\begin{CSLReferences}{1}{1}
\bibitem[\citeproctext]{ref-Atanasova2023-FaithfulnessTestsNatural}
Pepa Atanasova, Oana-Maria Camburu, Christina Lioma, Thomas Lukasiewicz,
Jakob Grue Simonsen, and Isabelle Augenstein. 2023.
\href{https://doi.org/10.18653/v1/2023.acl-short.25}{Faithfulness
{Tests} for {Natural Language Explanations}}. In \emph{Proc. 61st
{Annu}. {Meet}. {Assoc}. {Comput}. {Linguist}. {Vol}. 2 {Short Pap}.},
pages 283--294, Toronto, Canada. Association for Computational
Linguistics.

\bibitem[\citeproctext]{ref-Barbosa2025-EmpiricalAnalysisLarge}
André Barbosa, Igor Cataneo Silveira, and Denis Deratani Mauá. 2025.
\href{https://doi.org/10.5753/jbcs.2025.5817}{An {Empirical Analysis} of
{Large Language Models} for {Automated Cross-Prompt Essay Trait Scoring}
in {Brazilian Portuguese}}. \emph{J. Braz. Comput. Soc.},
31(1):857--870.

\bibitem[\citeproctext]{ref-Barrot2023-UsingChatGPTSecond}
Jessie S. Barrot. 2023.
\href{https://doi.org/10.1016/j.asw.2023.100745}{Using {ChatGPT} for
second language writing: {Pitfalls} and potentials}. \emph{Assess.
Writ.}, 57:100745.

\bibitem[\citeproctext]{ref-Bueno2026-ScoringExplanationsEvaluating}
Ivo Bueno, Babette Bühler, Philipp Stark, Tim Fütterer, Ulrich
Trautwein, Dorottya Demszky, Heather Hill, and Enkelejda Kasneci. 2026.
\href{https://doi.org/10.18653/v1/2026.findings-acl.375}{From {Scoring}
to {Explanations}: {Evaluating SHAP} and {LLM Rationales} for
{Rubric}-based {Teaching Quality Assessment}}. In \emph{Find. {Assoc}.
{Comput}. {Linguist}. {ACL} 2026}, pages 7590--7606, San Diego,
California, United States. Association for Computational Linguistics.

\bibitem[\citeproctext]{ref-Cao2020-DomainAdaptiveNeuralAutomated}
Yue Cao, Hanqi Jin, Xiaojun Wan, and Zhiwei Yu. 2020.
\href{https://doi.org/10.1145/3397271.3401037}{Domain-{Adaptive Neural
Automated Essay Scoring}}. In \emph{Proc. 43rd {Int}. {ACM SIGIR Conf}.
{Res}. {Dev}. {Inf}. {Retr}.}, pages 1011--1020, New York, NY, USA.
Association for Computing Machinery.

\bibitem[\citeproctext]{ref-Cho2024-DualscaleBERTUsing}
Minsoo Cho, Jin-Xia Huang, and Oh-Woog Kwon. 2024.
\href{https://doi.org/10.4218/etrij.2023-0324}{Dual-scale {BERT} using
multi-trait representations for holistic and trait-specific essay
grading}. \emph{ETRI J.}, 46(1):82--95.

\bibitem[\citeproctext]{ref-Chu2025-RationaleEssayScores}
SeongYeub Chu, Jong Woo Kim, Bryan Wong, and Mun Yong Yi. 2025.
\href{https://doi.org/10.18653/v1/2025.findings-naacl.322}{Rationale
{Behind Essay Scores}: {Enhancing S-LLM}'s {Multi-Trait Essay Scoring}
with {Rationale Generated} by {LLMs}}. In \emph{Find. {Assoc}. {Comput}.
{Linguist}. {NAACL} 2025}, pages 5811--5829, Albuquerque, New Mexico.
Association for Computational Linguistics.

\bibitem[\citeproctext]{ref-Chuang2025-LanguageAssessmentEra}
Ping-Lin Chuang and Xun Yan. 2025.
\href{https://doi.org/10.1016/j.system.2025.103846}{Language assessment
in the era of generative artificial intelligence: {Opportunities},
challenges, and future directions}. \emph{System}, 134:103846.

\bibitem[\citeproctext]{ref-Chuang2026-FaithLMFaithfulExplanations}
Yu-Neng Chuang, Guanchu Wang, Chia-Yuan Chang, Ruixiang Tang, Shaochen
Zhong, Fan Yang, Andrew Wen, Mengnan Du, Xuanting Cai, Vladimir
Braverman, and Xia Hu. 2026.
\href{https://doi.org/10.18653/v1/2026.eacl-long.177}{{FaithLM}:
{Towards Faithful Explanations} for {Large Language Models}}. In
\emph{Proc. 19th {Conf}. {Eur}. {Chapter Assoc}. {Comput}. {Linguist}.
{Vol}. 1 {Long Pap}.}, pages 3802--3824, Rabat, Morocco. Association for
Computational Linguistics.

\bibitem[\citeproctext]{ref-Demircan2024-SparseAutoencodersReveal}
Can Demircan, Tankred Saanum, Akshay K. Jagadish, Marcel Binz, and Eric
Schulz. 2024. \href{https://doi.org/10.48550/arXiv.2410.01280}{Sparse
{Autoencoders Reveal Temporal Difference Learning} in {Large Language
Models}}.

\bibitem[\citeproctext]{ref-Ding2025-ChatGPTAutomatedWriting}
Linqian Ding, Di Zou, and Lucas Kohnke. 2025.
\href{https://doi.org/10.1007/s10639-025-13775-3}{{ChatGPT} as an
automated writing evaluation tool: How students perceive it and how it
affects their writing}. \emph{Educ. Inf. Technol.}, 30(18):26777--26799.

\bibitem[\citeproctext]{ref-Elazar2021-AmnesicProbingBehavioral}
Yanai Elazar, Shauli Ravfogel, Alon Jacovi, and Yoav Goldberg. 2021.
\href{https://doi.org/10.1162/tacl_a_00359}{Amnesic {Probing}:
{Behavioral Explanation} with {Amnesic Counterfactuals}}. \emph{Trans.
Assoc. Comput. Linguist.}, 9:160--175.

\bibitem[\citeproctext]{ref-EngelhardJr1994-ExaminingRaterErrors}
George Engelhard Jr. 1994.
\href{https://doi.org/10.1111/j.1745-3984.1994.tb00436.x}{Examining
{Rater Errors} in the {Assessment} of {Written Composition With} a
{Many-Faceted Rasch Model}}. \emph{J. Educ. Meas.}, 31(2):93--112.

\bibitem[\citeproctext]{ref-Ferrando2023-ExplainingHowTransformers}
Javier Ferrando, Gerard I. Gállego, Ioannis Tsiamas, and Marta R.
Costa-jussà. 2023.
\href{https://doi.org/10.18653/v1/2023.acl-long.301}{Explaining {How
Transformers Use Context} to {Build Predictions}}. In \emph{Proc. 61st
{Annu}. {Meet}. {Assoc}. {Comput}. {Linguist}. {Vol}. 1 {Long Pap}.},
pages 5486--5513, Toronto, Canada. Association for Computational
Linguistics.

\bibitem[\citeproctext]{ref-Fiacco2023-ExtractingUnderstandingImplicit}
James Fiacco, David Adamson, and Carolyn Rose. 2023.
\href{https://doi.org/10.18653/v1/2023.bea-1.20}{Towards {Extracting}
and {Understanding} the {Implicit Rubrics} of {Transformer Based
Automatic Essay Scoring Models}}. In \emph{Proc. 18th {Workshop Innov}.
{Use NLP Build}. {Educ}. {Appl}. {BEA} 2023}, pages 232--241, Toronto,
Canada. Association for Computational Linguistics.

\bibitem[\citeproctext]{ref-Fisicaro1988-ReexaminationRelationHalo}
Sebastiano A. Fisicaro. 1988.
\href{https://doi.org/10.1037/0021-9010.73.2.239}{A reexamination of the
relation between halo error and accuracy}. \emph{J. Appl. Psychol.},
73(2):239--244.

\bibitem[\citeproctext]{ref-Gaggioli2025-AssessingReliabilityValidity}
Andrea Gaggioli, Giuseppe Casaburi, Leonardo Ercolani, Francesco
Collova', Pietro Torre, and Fabrizio Davide. 2025. Assessing the
{Reliability} and {Validity} of {Large Language Models} for {Automated
Assessment} of {Student Essays} in {Higher Education}.

\bibitem[\citeproctext]{ref-Hewitt2019-DesigningInterpretingProbes}
John Hewitt and Percy Liang. 2019.
\href{https://doi.org/10.18653/v1/D19-1275}{Designing and {Interpreting
Probes} with {Control Tasks}}. In \emph{Proc. 2019 {Conf}. {Empir}.
{Methods Nat}. {Lang}. {Process}. 9th {Int}. {Jt}. {Conf}. {Nat}.
{Lang}. {Process}. {EMNLP-IJCNLP}}, pages 2733--2743, Hong Kong, China.
Association for Computational Linguistics.

\bibitem[\citeproctext]{ref-Jacovi2021-AligningFaithfulInterpretations}
Alon Jacovi and Yoav Goldberg. 2021.
\href{https://doi.org/10.1162/tacl_a_00367}{Aligning {Faithful
Interpretations} with their {Social Attribution}}. \emph{Trans. Assoc.
Comput. Linguist.}, 9:294--310.

\bibitem[\citeproctext]{ref-Jiao2025-ExploringUtilitiesRationales}
Hong Jiao, Hanna Choi, and Haowei Hua. 2025.
\href{https://doi.org/10.48550/ARXIV.2510.27131}{Exploring the
{Utilities} of the {Rationales} from {Large Language Models} to {Enhance
Automated Essay Scoring}}.

\bibitem[\citeproctext]{ref-Johnsi2026-ComprehensiveSurveyMethodologies}
R. Johnsi and G. Bharadwaja Kumar. 2026.
\href{https://doi.org/10.1007/978-981-95-5891-9_47}{A {Comprehensive
Survey} on {Methodologies} for {Automatic Essay Scoring}}. In
\emph{Nonlinear {Complex Syst}. {Comput}. {Intell}. {Their Appl}.},
pages 677--725, Singapore. Springer Nature.

\bibitem[\citeproctext]{ref-Kabra2022-EvaluationToolkitRobustness}
Anubha Kabra, Mehar Bhatia, Yaman Kumar Singla, Junyi Jessy Li, and
Rajiv Ratn Shah. 2022.
\href{https://doi.org/10.1145/3493700.3493765}{Evaluation {Toolkit For
Robustness Testing Of Automatic Essay Scoring Systems}}. In \emph{Proc.
5th {Jt}. {Int}. {Conf}. {Data Sci}. {Manag}. {Data} 9th {ACM IKDD CODS}
27th {COMAD}}, pages 90--99, New York, NY, USA. Association for
Computing Machinery.

\bibitem[\citeproctext]{ref-Kucia2026-LLMEssayScoring}
Filip J. Kucia, Anirban Chakraborty, and Anna Wróblewska. 2026.
\href{https://doi.org/10.1007/978-3-032-29918-5_38}{{LLM Essay Scoring
Under Holistic} and~{Analytic Rubrics}: {Prompt Effects} and~{Bias}}. In
\emph{Comput. {Sci}. -- {ICCS} 2026 {Workshop}}, pages 531--546, Cham.
Springer Nature Switzerland.

\bibitem[\citeproctext]{ref-Madnani2026-ScoringWritingSpeaking}
Nitin Madnani, Phoebe Mulcaire, and Danwei Cai. 2026. Scoring {Writing}
\& {Speaking} on the {Duolingo English Test}.

\bibitem[\citeproctext]{ref-Mansour2024-CanLargeLanguage}
Watheq Ahmad Mansour, Salam Albatarni, Sohaila Eltanbouly, and Tamer
Elsayed. 2024. Can {Large Language Models Automatically Score
Proficiency} of {Written Essays}? In \emph{Proc. 2024 {Jt}. {Int}.
{Conf}. {Comput}. {Linguist}. {Lang}. {Resour}. {Eval}. {LREC-COLING}
2024}, pages 2777--2786, Torino, Italia. {ELRA and ICCL}.

\bibitem[\citeproctext]{ref-Mathew2026-LLMsNotGrade}
Jerin George Mathew, Sumayya Taher, Anindita Kundu, and Denilson
Barbosa. 2026. \href{https://doi.org/10.48550/ARXIV.2603.23714}{{LLMs Do
Not Grade Essays Like Humans}}.

\bibitem[\citeproctext]{ref-Mizumoto2023-ExploringPotentialUsing}
Atsushi Mizumoto and Masaki Eguchi. 2023.
\href{https://doi.org/10.1016/j.rmal.2023.100050}{Exploring the
potential of using an {AI} language model for automated essay scoring}.
\emph{Res. Methods Appl. Linguist.}, 2(2):100050.

\bibitem[\citeproctext]{ref-Myford2004-DetectingMeasuringRater}
Carol M. Myford and Edward W. Wolfe. 2004. Detecting and {Measuring
Rater Effects Using Many-Facet Rasch Measurement}: {Part II}. \emph{J.
Appl. Meas.}, 5(2):189--227.

\bibitem[\citeproctext]{ref-Pack2024-LargeLanguageModels}
Austin Pack, Alex Barrett, and Juan Escalante. 2024.
\href{https://doi.org/10.1016/j.caeai.2024.100234}{Large language models
and automated essay scoring of {English} language learner writing:
{Insights} into validity and reliability}. \emph{Comput. Educ. Artif.
Intell.}, 6:100234.

\bibitem[\citeproctext]{ref-Ramesh2022-AnAutomatedEssayScoringSystem}
Dadi Ramesh and Suresh Kumar Sanampudi. 2022.
\href{https://doi.org/10.1007/s10462-021-10068-2}{An automated essay
scoring systems: A systematic literature review}. \emph{Artif. Intell.
Rev.}, 55(3):2495--2527.

\bibitem[\citeproctext]{ref-Settles2020-MachineLearningDriven}
Burr Settles, Geoffrey T. LaFlair, and Masato Hagiwara. 2020.
\href{https://doi.org/10.1162/tacl_a_00310}{Machine {Learning}--{Driven
Language Assessment}}. \emph{Trans. Assoc. Comput. Linguist.},
8:247--263.

\bibitem[\citeproctext]{ref-Song2025-ExploringLLMAutoscoringb}
Dan Song, Won-Chan Lee, and Hong Jiao. 2025.
\href{https://doi.org/10.48550/arXiv.2507.19980}{Exploring {LLM
Autoscoring Reliability} in {Large-Scale Writing Assessments Using
Generalizability Theory}}.

\bibitem[\citeproctext]{ref-Wang2026-OpeningBlackboxLLMbased}
Manru Wang, Yihan Chen, Xiaoting Huang, and Yuxuan Lai. 2026.
\href{https://doi.org/10.1016/j.caeai.2026.100568}{Opening the blackbox
of {LLM}-based automated essay scoring: {Insights} into feature
weighting patterns and score validity}. \emph{Comput. Educ. Artif.
Intell.}, 10:100568.

\bibitem[\citeproctext]{ref-Wang2022-UseBertAutomated}
Yongjie Wang, Chuang Wang, Ruobing Li, and Hui Lin. 2022.
\href{https://doi.org/10.18653/v1/2022.naacl-main.249}{On the {Use} of
{Bert} for {Automated Essay Scoring}: {Joint Learning} of {Multi-Scale
Essay Representation}}. In \emph{Proc. 2022 {Conf}. {North Am}. {Chapter
Assoc}. {Comput}. {Linguist}. {Hum}. {Lang}. {Technol}.}, pages
3416--3425, Seattle, United States. Association for Computational
Linguistics.

\bibitem[\citeproctext]{ref-Wang2025-EvaluatingLargeLanguage}
Yuehan Wang, Jinyan Huang, Lun Du, Yuxin Guo, Ying Liu, and Rong Wang.
2025a. \href{https://doi.org/10.1016/j.caeai.2025.100481}{Evaluating
large language models as raters in large-scale writing assessments: {A}
psychometric framework for reliability and validity}. \emph{Comput.
Educ. Artif. Intell.}, 9:100481.

\bibitem[\citeproctext]{ref-Wang2025-LLMsCanPerform}
Zhengxiang Wang, Veronika Makarova, Zhi Li, Jordan Kodner, and Owen
Rambow. 2025b.
\href{https://doi.org/10.18653/v1/2025.acl-long.423}{{LLMs} can {Perform
Multi-Dimensional Analytic Writing Assessments}: {A Case Study} of {L2
Graduate-Level Academic English Writing}}. In \emph{Proc. 63rd {Annu}.
{Meet}. {Assoc}. {Comput}. {Linguist}. {Vol}. 1 {Long Pap}.}, pages
8637--8663, Vienna, Austria. Association for Computational Linguistics.

\bibitem[\citeproctext]{ref-Wu2023-CausalProxyModels}
Zhengxuan Wu, Karel D'Oosterlinck, Atticus Geiger, Amir Zur, and
Christopher Potts. 2023. Causal {Proxy Models} for {Concept}-based
{Model Explanations}. In \emph{Proc. 40th {Int}. {Conf}. {Mach}.
{Learn}.}, pages 37313--37334. PMLR.

\bibitem[\citeproctext]{ref-Xue2021-HierarchicalBERTBasedTransfer}
Jin Xue, Xiaoyi Tang, and Liyan Zheng. 2021.
\href{https://doi.org/10.1109/ACCESS.2021.3110683}{A {Hierarchical
BERT-Based Transfer Learning Approach} for {Multi-Dimensional Essay
Scoring}}. \emph{IEEE Access}, 9:125403--125415.

\bibitem[\citeproctext]{ref-Yamashita2025-ExploringPotentialBiases}
Taichi Yamashita. 2025.
\href{https://doi.org/10.1177/02655322251329435}{Exploring potential
biases in {GPT-4}o's ratings of {English} language learners' essays}.
\emph{Lang. Test.}, 42(3):344--358.

\bibitem[\citeproctext]{ref-Yancey2023-RatingShortL2}
Kevin P. Yancey, Geoffrey Laflair, Anthony Verardi, and Jill Burstein.
2023. \href{https://doi.org/10.18653/v1/2023.bea-1.49}{Rating {Short L2
Essays} on the {CEFR Scale} with {GPT-4}}. In \emph{Proc. 18th {Workshop
Innov}. {Use NLP Build}. {Educ}. {Appl}. {BEA} 2023}, pages 576--584,
Toronto, Canada. Association for Computational Linguistics.

\bibitem[\citeproctext]{ref-Yavuz2025-UtilizingLargeLanguage}
Fatih Yavuz, Özgür Çelik, and Gamze Yavaş Çelik. 2025.
\href{https://doi.org/10.1111/bjet.13494}{Utilizing large language
models for {EFL} essay grading: {An} examination of reliability and
validity in rubric-based assessments}. \emph{Br. J. Educ. Technol.},
56(1):150--166.

\bibitem[\citeproctext]{ref-Zhang2023-BestPracticesActivation}
Fred Zhang and Neel Nanda. 2023. Towards {Best Practices} of {Activation
Patching} in {Language Models}: {Metrics} and {Methods}.

\end{CSLReferences}
\end{document}